\documentclass[11pt,a4paper]{article}

\usepackage[utf8]{inputenc}
\usepackage{fontenc}
\usepackage{amsmath}
\usepackage{amsfonts}
\usepackage{amssymb}
\usepackage{graphicx}
\usepackage{booktabs}
\usepackage{multicol}
\usepackage[margin=1in]{geometry}
\usepackage{hyperref}
\usepackage{url}
\usepackage{authblk}
\usepackage{nicefrac}
\usepackage{microtype}
\usepackage{doi}
\usepackage{caption}
\usepackage{subcaption}
\usepackage{latexsym}
\usepackage{inconsolata}
\usepackage{times}

\hypersetup{
    colorlinks=true,
    linkcolor=blue,
    filecolor=magenta,
    urlcolor=cyan,
    pdftitle={TensorBLEU},
    pdfpagemode=FullScreen,
}

\title{
    \textbf{TensorBLEU: Vectorized GPU-based BLEU Score Implementation for Per-Sentence In-Training Evaluation}
}
\author{Adam Filipek (adamfilipek@rxai.dev)}
\affil{Reactive AI (https://rxai.dev)}
\date{October 2025}

\begin{document}
\maketitle
\begin{abstract}
Modern natural language processing models have achieved unprecedented scale, yet the tools for their evaluation often remain a computational bottleneck, limiting the pace of research. This is particularly acute for in-training evaluation metrics, such as per-sentence reward signals in Reinforcement Learning, which must operate efficiently on batches of token IDs directly on the GPU. In this paper, we introduce TensorBLEU, a novel implementation of the BLEU metric designed from the ground up for this specific use case. Our approach is fully vectorized for GPU-accelerated, per-sentence computation within PyTorch and introduces a memory-efficient counting mechanism. By creating a compact, batch-specific dictionary of n-grams using \texttt{torch.unique}, our method avoids the prohibitive memory costs of traditional hashing-based vectorization, making it practical for large-vocabulary models. We benchmark TensorBLEU against NLTK, the standard library for token-ID-based BLEU calculation on the CPU. Experiments show that TensorBLEU provides speedups of over 13x on consumer-grade GPUs (NVIDIA T4) and exceeding 40x on data-center-class hardware (NVIDIA A100). This performance transforms a significant bottleneck into a negligible part of the training loop. By clearly defining its role as a "Token-ID BLEU" for development purposes and open-sourcing our implementation, we provide a powerful tool for accelerating research in areas like RL-based model fine-tuning.
\end{abstract}

\section{Introduction}

The advancement in Natural Language Processing (NLP) has been driven by the scaling of neural architectures and datasets. However, while models have become exponentially more powerful, the tools for their evaluation have often failed to keep pace. Metrics that cannot efficiently process batches of data in parallel on the GPU create an unnecessary bottleneck related to data transfer and sequential computation.

This problem is most critical for applications that require a metric to be computed repeatedly inside the training loop. A prime example is Reinforcement Learning (RL) for fine-tuning language models, where a dense reward signal is needed for each generated sample in a batch. A single "corpus" score for the entire batch provides a weak, averaged signal, whereas a per-sample reward is necessary for effective policy gradient updates. A slow, CPU-bound metric can dominate the computation time, making such approaches impractical. The BLEU score, a standard for assessing text generation quality, is a desirable candidate for such a reward, but traditional implementations are ill-suited for this role \cite{papineni2002bleu}. While libraries like NLTK can compute BLEU from integer token IDs, they require moving tensors from GPU to CPU, converting them to lists, and processing each sample in a Python loop, creating a severe performance bottleneck. On the other hand, more efficient SacreBLEU \cite{post2018call} requires inputs in text format, so in case of RL generated tokens have to be decoded, only for reward calculation, what is another potential bottleneck.

In our case, the need to implement an efficient, GPU-based BLEU calculation arose during training of our Reactive Transformer models \cite{rxt2025}. During the memory reinforcement learning stage, BLEU calculation on the CPU was combined with cosine similarity calculation on the GPU, which required continuous data copying between devices.

\paragraph{Our Contribution: TensorBLEU} In this paper, we present TensorBLEU, a re-architected implementation of the BLEU metric designed specifically for batched, per-sentence, vectorized computation on token IDs within the PyTorch environment. Our work makes two key contributions:
\begin{enumerate}
    \item \textbf{A Memory-Efficient Vectorized Algorithm for Per-Sentence BLEU:} We introduce a novel method for n-gram counting that avoids the memory explosion of naive vectorization. Instead of using a large hash space, we use \texttt{torch.unique} on the n-gram tensors themselves to create a compact, batch-specific dictionary. This allows for efficient, parallel per-sentence counting via a "batched bincount" technique in a memory space proportional to the number of unique n-grams in the batch, not the vocabulary size.
    \item \textbf{A High-Performance In-Training Metric with Demonstrated Scalability:} We benchmark TensorBLEU on both consumer (NVIDIA T4) and data-center (NVIDIA A100) GPUs, demonstrating speedups that scale with hardware capabilities and effectively remove the evaluation bottleneck across different research environments.
\end{enumerate}

\section{Background and Related Work}

\subsection{The Original BLEU Metric}
The BLEU (Bilingual Evaluation Understudy) metric was introduced by Papineni et al. (2002) to address the slow and expensive process of human evaluation for machine translation \cite{papineni2002bleu}. It is based on two components: modified n-gram precision and a brevity penalty. Modified n-gram precision uses a "clipping" mechanism to prevent systems from over-generating common, correct words. The brevity penalty (BP) penalizes candidate translations that are shorter than their references. The final score is the geometric mean of the precisions (typically for n=1 to 4), multiplied by the BP:
$$\text{BLEU} = \text{BP} \times \exp\left(\sum_{n=1}^{N} w_n \log p_n\right)$$

\subsection{The Need for Standardization: SacreBLEU}
As BLEU became the dominant metric, a reproducibility crisis emerged. Scores varied wildly between papers due to undisclosed differences in preprocessing and tokenization, making fair comparison impossible \cite{post2018call}. To solve this, Post (2018) introduced SacreBLEU, a standardized implementation that manages the entire evaluation pipeline, including a canonical tokenization scheme. This ensures that reported scores are comparable and reproducible, establishing SacreBLEU as the gold standard for final, publication-ready evaluation \cite{post2018call}.

\subsection{From Evaluation to Optimization: BLEU as an RL Reward}
Standard training of sequence models via word-level cross-entropy suffers from "exposure bias": the model is only trained on ground-truth prefixes, not its own, often imperfect, predictions \cite{ranzato2015sequence}. A solution is to train at the sequence level by directly optimizing a metric like BLEU. Since BLEU is non-differentiable, Reinforcement Learning (RL) provides a framework for this optimization. The model acts as a policy, generation is a sequence of actions, and the final BLEU score serves as the reward.

The seminal work by Ranzato et al. (2015) introduced this sequence-level training paradigm to NLP, using the REINFORCE algorithm to directly optimize for metrics like BLEU \cite{ranzato2015sequence}. Subsequent work, such as Li et al. (2016), successfully applied this technique to dialogue generation, demonstrating its broader utility \cite{li2016deep}. However, the computational cost of calculating the BLEU reward for every batch on the CPU has remained a major barrier to the widespread adoption of this powerful technique. TensorBLEU is designed to remove this barrier.

\subsection{Token-ID BLEU vs. Linguistic BLEU}
It is crucial to distinguish between two modes of BLEU calculation.
\paragraph{Linguistic BLEU} is the standard for final model evaluation (e.g., SacreBLEU). It operates on detokenized text and applies its own standardized tokenization to ensure scores are reproducible.
\paragraph{Token-ID BLEU} in contrast, operates directly on the integer outputs of a model's tokenizer. The n-grams are sequences of subword IDs. While unsuitable for final reporting, this metric is perfectly suited for \textit{internal, relative} evaluation during the development cycle (e.g., as an RL reward), where the tokenizer is held constant. TensorBLEU is a high-performance implementation of Token-ID BLEU.

\section{TensorBLEU: A Memory-Efficient Vectorized Implementation}
This section details the core algorithm of TensorBLEU, which is designed to compute a separate BLEU score for each candidate-reference pair in a batch in a fully vectorized manner, without resorting to Python loops.

\subsection{Vectorized n-gram Extraction with \texttt{unfold}}
The first step is to extract all n-grams from the entire batch of sentences in parallel. We use the \texttt{Tensor.unfold} method in PyTorch. For a batch of token sequences with shape \texttt{(batch\_size, seq\_len)}, applying \texttt{tensor.unfold(dimension=1, size=n, step=1)} returns a view of the original tensor containing all n-gram slices, with a shape of \texttt{(batch\_size, num\_ngrams, n)}. This operation is highly efficient as it avoids data copying and processes all sentences simultaneously.

\subsection{Memory-Efficient Counting via a Unified n-gram Dictionary}
A naive approach to vectorizing n-gram counting involves hashing each n-gram slice into a unique integer and using a counting tensor of size $(batch\_size, V^n)$, where V is the vocabulary size. This leads to a memory explosion for modern vocabularies.

To solve this, we developed a memory-efficient method that operates in a compact space. The algorithm proceeds as follows:
\begin{enumerate}
    \item \textbf{Unified N-gram Collection:} For a given order $n$, we extract all valid n-grams from both the candidate and all reference sentences in the batch and flatten them into a single large tensor of shape \texttt{(total\_ngrams, n)}.
    \item \textbf{Compact Dictionary Creation:} We apply \texttt{torch.unique(all\_ngrams, dim=0, \newline return\_inverse=True)}. This is the key step. It returns two tensors:
    \begin{itemize}
        \item \texttt{unique\_ngrams}: A tensor containing only the unique n-grams that actually appear in the current batch.
        \item \texttt{inverse\_indices}: A 1D tensor mapping each original n-gram to its new, compact ID (i.e., its index in \texttt{unique\_ngrams}).
    \end{itemize}
    This step effectively creates a batch-specific "dictionary" of n-grams, where the memory required is proportional to the number of unique n-grams present, not the theoretical maximum.
\end{enumerate}

\subsection{Batched Counting and Clipping with Offset Bincounting}
With the compact IDs from the previous step, we can now perform counting for each sentence in parallel. We use a novel "batched bincount" technique.
\begin{enumerate}
    \item \textbf{The Offset Mechanism:} For each sentence $i$ in the batch, we add a unique offset, calculated as $i \times \texttt{num\_unique\_ngrams}$, to its compact n-gram IDs. This ensures that the IDs for each sentence occupy a unique, non-overlapping range.
    \item \textbf{Single Bincount Operation:} These offset IDs are then flattened into a single 1D tensor. A single call to \texttt{torch.bincount} on this tensor computes the n-gram counts for all sentences simultaneously. The resulting flat tensor is then reshaped to \texttt{(batch\_size, num\_unique\_ngrams)}.
    \item \textbf{Reference Counts and Clipping:} The same process is applied to the reference n-grams. To obtain the final reference counts for clipping, a \texttt{torch.maximum} operation is taken over the count tensors derived from each reference set. The final clipping is a simple, vectorized \texttt{torch.minimum(candidate\_counts, reference\_max\_counts)}.
\end{enumerate}

\subsection{Final Score Aggregation and The \texttt{tensor\_corpus\_bleu} Variant}
With the clipped counts (numerators) and total candidate n-gram counts (denominators) computed for each sentence and each n-gram order, the rest of the calculation is performed element-wise across the batch dimension. The modified precisions $p_n$, brevity penalty, and final geometric mean are assembled using standard PyTorch functions. Our implementation also includes standard smoothing methods ('floor', 'add-k', 'exp') as described by Chen and Cherry (2014) \cite{chen2014systematic}.

In addition to the per-sentence function, we provide a \texttt{tensor\_corpus\_bleu} variant that computes a single score for the entire batch by aggregating statistics before calculating precision. The performance of this variant is nearly identical to the per-sentence version. This is because the most computationally expensive steps—n-gram extraction with \texttt{unfold} and the creation of the compact dictionary with \texttt{torch.unique}—are performed on the entire batch's n-grams in both cases and dominate the runtime. The additional complexity in the per-sentence version (offset calculation, per-sentence aggregation) consists of lightweight arithmetic operations that are negligible on a GPU. This allows researchers to obtain a more granular, per-sentence reward signal at virtually no performance penalty compared to a less useful corpus-level score.

\section{Experimental Design}

\subsection{Correctness Verification}
To prove that TensorBLEU correctly implements the BLEU algorithm for token IDs, we verified that it produces numerically identical results to NLTK's \texttt{sentence\_bleu} function when given the same lists of token IDs, weights, and smoothing function. Our implementation consistently achieved equivalence within the margin of floating-point precision ($< 10^{-6}$).

\subsection{Performance Benchmarking}
\paragraph{Objective} To quantify the speedup of TensorBLEU over the standard CPU-based method for calculating Token-ID BLEU.
\paragraph{Reference Implementation} We compare against NLTK's \texttt{sentence\_bleu}, which is the standard library for this task. The NLTK implementation is run on the CPU and involves iterating through the batch in a Python loop.
\paragraph{Hardware and Data} Experiments were conducted across two distinct hardware tiers:
\begin{itemize}
    \item \textbf{Consumer-Grade:} A Google Colab environment featuring an NVIDIA T4 GPU (16GB VRAM) and an Intel Xeon CPU (2 cores, 13GB RAM).
    \item \textbf{Data-Center-Grade:} A Novita.ai cloud instance featuring an NVIDIA A100 GPU (80GB HBM2e VRAM) and a 14 vCPU instance (240GB RAM).
\end{itemize}
We used batches of token sequences of two lengths: 256 tokens (typical for many tasks) and 1024 tokens (representing longer-form generation). All computations for TensorBLEU were performed on the GPU using float32 precision.
\paragraph{Variables} We measured the wall-clock execution time (mean of 5 runs) while varying the batch size, using values of 16, 32, 64, 128, 256, and 512.

\section{Results and Analysis}

\subsection{Performance Results}
The performance results are presented in Table \ref{tab:t4_performance} for the consumer-grade NVIDIA T4 GPU and Table \ref{tab:a100_performance} for the data-center-grade NVIDIA A100 GPU.

\begin{table}[h!]
\centering
\caption{Performance on NVIDIA T4 GPU (mean execution time in seconds). Lower is better.}
\label{tab:t4_performance}
\begin{tabular}{lrrr}
\toprule
\textbf{Batch Size} & \textbf{NLTK (CPU)} & \textbf{TensorBLEU (GPU)} & \textbf{Speedup Factor} \\
\midrule
\multicolumn{4}{c}{\textit{Sequence Length: 256 tokens}} \\
\midrule
32 & 0.042s & 0.011s & 3.8x \\
64 & 0.079s & 0.015s & 5.3x \\
128 & 0.163s & 0.016s & 10.2x \\
256 & 0.333s & 0.035s & 9.5x \\
512 & 0.702s & 0.085s & 8.3x \\
\midrule
\multicolumn{4}{c}{\textit{Sequence Length: 1024 tokens}} \\
\midrule
16 & 0.072s & 0.011s & 6.5x \\
32 & 0.131s & 0.015s & 8.7x \\
64 & 0.252s & 0.019s & 13.3x \\
128 & 0.482s & 0.036s & 13.4x \\
256 & 0.974s & 0.084s & 11.6x \\
\bottomrule
\end{tabular}
\end{table}

\begin{table}[h!]
\centering
\caption{Performance on NVIDIA A100 GPU for 1024-token sequences. Lower is better.}
\label{tab:a100_performance}
\begin{tabular}{lrrr}
\toprule
\textbf{Batch Size} & \textbf{NLTK (CPU)} & \textbf{TensorBLEU (GPU)} & \textbf{Speedup Factor} \\
\midrule
32 & 0.107s & 0.009s & 11.9x \\
64 & 0.200s & 0.010s & 20.0x \\
128 & 0.380s & 0.013s & 29.2x \\
256 & 0.764s & 0.019s & 40.2x \\
512 & 1.525s & 0.041s & 37.2x \\
\bottomrule
\end{tabular}
\end{table}

\begin{figure}
    \centering
    \includegraphics[width=0.75\linewidth]{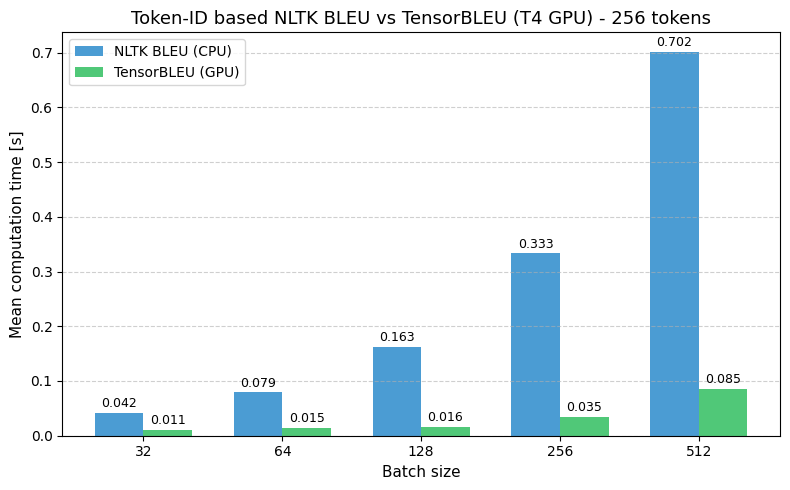}
    \caption{Tests on T4 GPU (Colab) with 256 tokens sentences}
    \label{fig:t4_small}
\end{figure}

\begin{figure}
    \centering
    \includegraphics[width=0.75\linewidth]{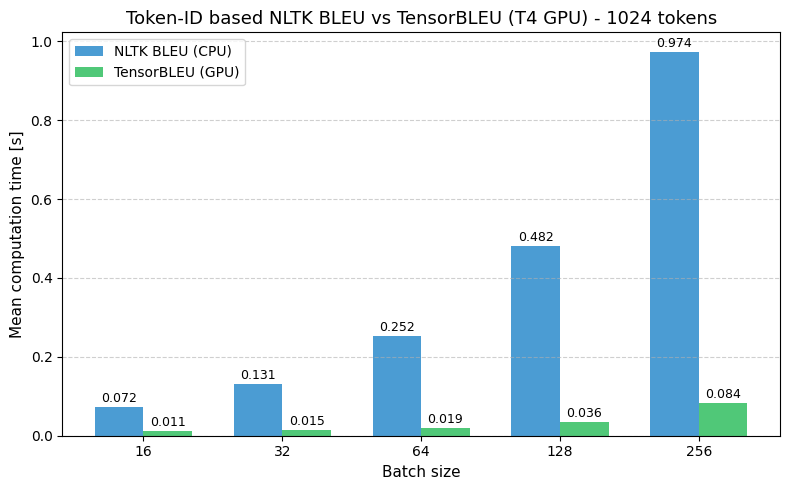}
    \caption{Tests on T4 GPU (Colab) with 1024 tokens sentences}
    \label{fig:t4_big}
\end{figure}

\begin{figure}
    \centering
    \includegraphics[width=0.75\linewidth]{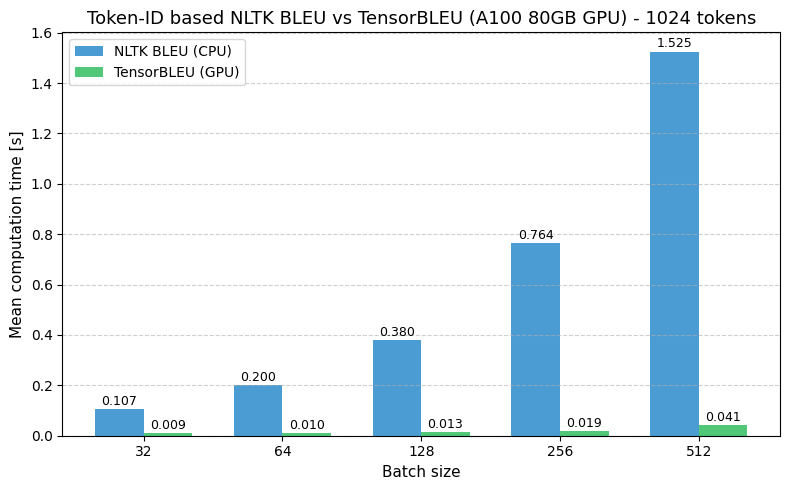}
    \caption{Tests on A100 80GB GPU (and stronger CPU) with 1024 tokens sentences}
    \label{fig:a100}
\end{figure}

\subsection{Analysis of Results}
The data clearly demonstrate the profound performance advantage of our vectorized, GPU-based approach. The NLTK implementation, being a serial loop over sentences, exhibits near-linear time complexity with respect to batch size. In contrast, TensorBLEU shows sub-linear scaling, as the fixed costs of launching GPU kernels are amortized over an increasing number of parallel computations.

\paragraph{Impact of Sequence Length} Comparing the results on the T4 GPU (Table \ref{tab:t4_performance}), we observe that the speedup advantage of TensorBLEU grows with sequence length. For a batch size of 128, the speedup increases from 10.2x for 256-token sequences to 13.4x for 1024-token sequences. This is because the number of n-grams to process grows significantly with sequence length, heavily penalizing the iterative CPU approach, while the parallel GPU architecture can absorb this increased workload much more efficiently.

\paragraph{Impact of Hardware Tier} The comparison between the T4 (Table \ref{tab:t4_performance}) and A100 (Table \ref{tab:a100_performance}) results for 1024-token sequences highlights the algorithm's scalability. On the A100, the speedup factor reaches a remarkable 40.2x at a batch size of 256. This demonstrates that the algorithm's design effectively leverages the superior memory bandwidth and computational power of high-end GPUs. The performance gains are super-linear, indicating that the implementation is not limited by its own logic but by the underlying hardware, which is a hallmark of a well-designed, scalable algorithm. The A100's massive memory bandwidth is particularly beneficial for the \texttt{torch.unique} operation on a very large tensor of n-grams.

\paragraph{The Bottleneck Vanishes} For a typical training scenario (e.g., batch size 256, 1024 tokens), TensorBLEU reduces the evaluation time from a significant fraction of a second (764ms on a capable CPU) to just 19ms on an A100 GPU. This effectively transforms the metric calculation from a potential training bottleneck into a negligible overhead.

\section{Discussion and Future Work}

\subsection{Implications for NLP Research}
The primary implication of our work is that it makes large-scale RL fine-tuning of language models using BLEU as a dense reward signal computationally cheap and practical. By reducing the computation time for a batch from hundreds of milliseconds to just a few, TensorBLEU removes a critical bottleneck that previously made such approaches prohibitively slow. This accelerates the research and development cycle, allowing for more extensive experimentation, hyperparameter sweeps, and application to larger models where every second of training time counts.

\subsection{Limitations and Proper Usage}
It is essential to use TensorBLEU correctly. As a "Token-ID BLEU" metric, its results are dependent on the tokenizer and are not directly comparable to scores from other models using different tokenizers. It is an \textbf{internal development and optimization tool}, ideal for measuring relative improvements during training. For final, publication-ready results that are comparable across the field, researchers must continue to use standardized, text-based tools like SacreBLEU \cite{post2018call}, which ensure consistent tokenization and processing.

\subsection{Future Work}
The core contribution of this paper extends beyond a fast BLEU implementation to a generalizable methodology for vectorizing n-gram-based metrics on GPUs using \texttt{torch.unique} as a memory-efficient hashing mechanism. Based on this, we propose several directions for future work:
\begin{enumerate}
    \item \textbf{Generalizing the Technique:} The vectorized counting methodology can be extended to other n-gram-based metrics like ROUGE and METEOR. Developing a suite of high-performance "TensorMetrics" would provide the community with a powerful toolkit for GPU-accelerated evaluation.
    \item \textbf{Integration with RL Libraries:} To facilitate adoption, we plan to develop official integrations and tutorials for popular RL libraries like Hugging Face's TRL and AllenAI's RL4LMs, making it trivial for researchers to use TensorBLEU as a reward function.
    \item \textbf{Exploring Further Optimizations:} We plan to investigate the performance impact of lower-precision data types like \texttt{bfloat16} and explore the use of custom CUDA kernels for the counting mechanism to potentially achieve even greater speedups on specific hardware.
\end{enumerate}

\section{Conclusion}
In this paper, we presented TensorBLEU, a fully vectorized, memory-efficient, and GPU-accelerated implementation of the BLEU metric designed for per-sentence, in-training evaluation on token IDs. By leveraging a novel counting mechanism based on \texttt{torch.unique}, our method is practical for large-vocabulary models and avoids the memory explosion of naive vectorization. It achieves speedups of over 13x on consumer-grade hardware and over 40x on data-center GPUs compared to the standard CPU-based NLTK implementation. This performance demonstrates excellent scalability and effectively eliminates the evaluation bottleneck for in-training use cases. By clearly defining its scope and open-sourcing the code, we provide the NLP community with a critical piece of infrastructure to accelerate research in computationally intensive paradigms like Reinforcement Learning.

\section*{Acknowledgements}
The article was created in cooperation with Google Gemini 2.5 Pro \cite{gemini2025} in "Deep Research" mode, which helped with the analysis of the algorithm and formatting the final version of the research paper.

\section*{Code and Documentation}
Implementation and usage documentation for TensorBLEU in both sentence and corpus modes is publicly available as "Free Component" in our RxLM framework (https://github.com/RxAI-dev/rxlm) as \texttt{rxlm.metrics.tensorbleu} module. "Free Components" in \textbf{Reactive AI Framework License (RAFL) v1.0} are available under \textbf{Apache-2.0} license terms.


\begin{thebibliography}{9}

\bibitem{papineni2002bleu}
Kishore Papineni, Salim Roukos, Todd Ward, and Wei-Jing Zhu.
\newblock Bleu: a method for automatic evaluation of machine translation.
\newblock In \emph{Proceedings of the 40th Annual Meeting of the Association for Computational Linguistics}, pages 311--318, 2002.

\bibitem{post2018call}
Matt Post.
\newblock A call for clarity in reporting bleu scores.
\newblock In \emph{Proceedings of the Third Conference on Machine Translation: Research Papers}, pages 186--191, 2018.

\bibitem{ranzato2015sequence}
Marc'Aurelio Ranzato, Sumit Chopra, Michael Auli, and Wojciech Zaremba.
\newblock Sequence level training with recurrent neural networks.
\newblock \emph{arXiv preprint arXiv:1511.06732}, 2015.

\bibitem{li2016deep}
Jiwei Li, Will Monroe, Alan Ritter, Dan Jurafsky, Michel Galley, and Jianfeng Gao.
\newblock Deep reinforcement learning for dialogue generation.
\newblock In \emph{Proceedings of the 2016 Conference on Empirical Methods in Natural Language Processing}, pages 1192--1202, 2016.

\bibitem{chen2014systematic}
Boxing Chen and Colin Cherry.
\newblock A systematic comparison of smoothing techniques for sentence-level bleu.
\newblock In \emph{Proceedings of the Ninth Workshop on Statistical Machine Translation}, pages 362--367, 2014.

\bibitem{rxt2025}
Adam Filipek.
\newblock Reactive Transformer (RxT) - Stateful Real-Time Processing for Event-Driven Reactive Language Models.
\newblock \emph{arXiv preprint arXiv:2510.03561}, 2025.

\bibitem{gemini2025}
Google DeepMind.
\newblock Gemini 2.5: Pushing the Frontier with Advanced Reasoning, Multimodality, Long Context, and Next Generation Agentic Capabilities.
\newblock \emph{arXiv preprint arXiv:2507.06261}, 2025.
	

\end{thebibliography}
\end{document}